\documentclass[10pt,a4paper]{article}

\usepackage[margin=0.72in]{geometry}
\usepackage{amsmath,amssymb,amsthm}
\usepackage{booktabs,array}
\usepackage{hyperref}
\usepackage{titlesec}
\usepackage{microtype}
\usepackage{xcolor}

\hypersetup{
    colorlinks=true,
    linkcolor=blue!50!black,
    citecolor=blue!50!black,
    urlcolor=blue!50!black
}

\titlespacing*{\section}{0pt}{7pt}{3pt}
\titlespacing*{\subsection}{0pt}{5pt}{2pt}
\titlespacing*{\subsubsection}{0pt}{4pt}{1pt}
\newtheorem{definition}{Definition}
\newtheorem{conjecture}{Conjecture}

\title{\vspace{-1.2em}\textbf{A CAP-like Trilemma for Large Language Models: Correctness, Non-bias, and Utility under Semantic Underdetermination}\vspace{-0.4em}}

\author{
Vinu Ellampallil Venugopal\\
International Institute of Information Technology Bangalore, Bangalore, India\\
ORCID: \href{https://orcid.org/0000-0003-4429-9932}{0000-0003-4429-9932}
}

\date{\vspace{-1.0em}}

\begin{document}
\maketitle
\vspace{-1.3em}

\begin{abstract}
The CAP theorem states that a distributed system cannot simultaneously guarantee consistency, availability, and partition tolerance under network partition. Inspired by this result, this paper formulates a CAP-like conjecture for Large Language Models (LLMs). The proposed trilemma states that, under semantic underdetermination, an LLM cannot always simultaneously guarantee strong correctness, strict non-bias, and high utility. A prompt is semantically underdetermined when the given premises do not determine a unique answer. In such cases, a useful and decisive response requires the model to introduce a selection criterion, preference, prior, or value ordering. If this criterion is not supplied by the user or justified by the available premises, the response becomes biased in a broad selection-theoretic sense. Conversely, if the model avoids unsupported preferences, it may preserve correctness and non-bias but may reduce utility through refusal, hedging, or clarification. The paper formalizes this correctness--non-bias--utility trilemma, develops examples, and argues that certain LLM failures arise not merely from model limitations but from the structure of underdetermined decision requests.
\end{abstract}


\section{Introduction}

The CAP theorem is a foundational result in distributed systems. It states that consistency, availability, and partition tolerance cannot all be guaranteed in the presence of a network partition \cite{brewer2000,gilbert2002}. When a partition occurs, a distributed system must either preserve consistency by refusing or delaying some requests, or preserve availability by responding despite the risk of inconsistency. This paper proposes that an analogous tension exists in LLM response generation. The stress condition in distributed systems is a network partition; the stress condition in LLMs is semantic underdetermination.

A prompt is semantically underdetermined when the facts, assumptions, and premises available in the prompt are insufficient to determine a unique answer. Many user queries are of this kind: ``Which candidate should be selected?'', ``Which city is better?'', ``What is the right decision?'', or ``Should a company prioritize profit or employee welfare?'' Such questions request a singleton answer, but the prompt itself often does not specify the decision criterion under which a singleton answer would be justified. LLM alignment research already recognizes that models may generate outputs that are untruthful, toxic, biased, or unhelpful, and that training with human feedback can improve helpfulness and alignment with user intent \cite{ouyang2022}. However, the present paper focuses on a more structural issue: in some cases, the user asks for a decision that the premises do not logically determine.

The central conjecture is that correctness, non-bias, and utility form a trilemma under semantic underdetermination. If the model prioritizes correctness and non-bias, it may refuse or ask for clarification, thereby reducing utility. If it prioritizes utility by giving a decisive answer, it must introduce some criterion or preference, which may introduce bias. If it attempts to remain neutral while still being useful, it may weaken correctness by presenting unsupported equivalence or balance.

\section{From CAP to LLM Response Generation}

In the classical CAP setting, consistency means that every read reflects the most recent write or returns an error, availability means that every request receives a non-error response, and partition tolerance means that the system continues operating despite network failures. Gilbert and Lynch formalized Brewer's conjecture and proved that, in an asynchronous network model, availability and atomic consistency cannot both be guaranteed under partition \cite{gilbert2002}. The standard logical form is:
\[
P \Rightarrow \neg(C \land A),
\]
where \(P\) denotes partition, \(C\) denotes consistency, and \(A\) denotes availability.

The proposed LLM analogue replaces network partition with semantic underdetermination. Let \(x\) be a prompt, \(F_x\) be the set of facts and premises explicitly available in \(x\), and \(\mathcal{A}\) be the answer space. The set of premise-compatible answers is:
\[
S_x = \{a \in \mathcal{A} \mid a \text{ is compatible with } F_x\}.
\]
If \(|S_x|=1\), the prompt has a singleton solution. If \(|S_x|>1\), the prompt is underdetermined. We write:
\[
D(x) \iff |S_x|>1.
\]
The conjectured analogue of CAP is:
\[
D(x) \Rightarrow \neg(C_{\mathrm{strong}} \land NB_{\mathrm{strict}} \land U_{\mathrm{decisive}}),
\]
where \(C_{\mathrm{strong}}\) denotes strong correctness, \(NB_{\mathrm{strict}}\) denotes strict non-bias, and \(U_{\mathrm{decisive}}\) denotes high utility through a concrete answer.

\section{Correctness, Non-bias, and Utility}

\begin{definition}[Strong correctness]
An output \(y\) is strongly correct for prompt \(x\) if it is entailed by the available premises:
\[
C_{\mathrm{strong}}(y,x)=1 \iff F_x \models y.
\]
This is a strict notion. It does not merely require plausibility; it requires that the conclusion follow from the premises.
\end{definition}

For factual prompts, strong correctness is often achievable. If \(x\) asks, ``What is the capital of France?'', then \(F_x\) plus external factual knowledge supports a singleton answer, namely Paris. However, for non-factual prompts, a unique answer may not be entailed. This distinction is related to truthfulness concerns in LLMs: models may generate answers that sound plausible but are not warranted, and benchmarks such as TruthfulQA explicitly measure whether models avoid false or misleading answers learned from imitation of human text \cite{lin2022}.

\begin{definition}[Strict non-bias]
An output \(y\) satisfies strict non-bias if the model does not privilege one candidate answer over another unless that preference is justified by \(F_x\). For \(a_i,a_j \in S_x\), if
\[
F_x \not\models a_i \succ a_j
\quad\text{and}\quad
F_x \not\models a_j \succ a_i,
\]
then the model should not present either answer as uniquely preferable without stating an additional assumption.
\end{definition}

Here, bias is used broadly. It includes harmful social bias, but it also includes any unsupported selection preference, prior, ranking rule, or normative criterion. This broader sense is compatible with empirical findings that language representations can encode human-like biases from ordinary language corpora \cite{caliskan2017}, but the present paper is not limited to demographic or social bias. It treats bias as a selection mechanism introduced when the premises alone do not determine a choice.

\begin{definition}[Utility]
Utility measures how useful, informative, responsive, and actionable a response is. A simple decomposition is:
\[
U(y,x)=f(R(y,x),I(y,x),A(y,x),D_s(y,x)),
\]
where \(R\) is relevance, \(I\) is informativeness, \(A\) is actionability, and \(D_s\) is decisiveness. Decisive utility, written \(U_{\mathrm{decisive}}\), is the special case in which the response gives a concrete answer rather than only asking for clarification or refusing.
\end{definition}

The helpfulness dimension of LLM alignment is often evaluated together with truthfulness and harmlessness \cite{ouyang2022}. The present conjecture suggests that these desiderata are not always independently maximizable: in underdetermined prompts, helpfulness can require selection, and selection can require assumptions.

\section{The Correctness--Non-bias--Utility Trilemma}

\begin{conjecture}[Correctness--non-bias--utility trilemma]
For a semantically underdetermined prompt \(x\), an LLM cannot always simultaneously satisfy strong correctness, strict non-bias, and decisive utility:
\[
D(x) \Rightarrow \neg(C_{\mathrm{strong}} \land NB_{\mathrm{strict}} \land U_{\mathrm{decisive}}).
\]
\end{conjecture}

The trilemma has three pairwise forms:
\[
C_{\mathrm{strong}} \land NB_{\mathrm{strict}} \Rightarrow \neg U_{\mathrm{decisive}},
\]
\[
NB_{\mathrm{strict}} \land U_{\mathrm{decisive}} \Rightarrow \neg C_{\mathrm{strong}},
\]
\[
C_{\mathrm{strong}} \land U_{\mathrm{decisive}} \Rightarrow \neg NB_{\mathrm{strict}}.
\]
The first relation states that if the model refuses to introduce unsupported assumptions, it may lose decisiveness. The second states that if the model remains neutral while still trying to be decisive, it may make claims not entailed by the premises. The third states that if the model gives a correct answer under some criterion and remains useful, the criterion itself must be supplied; if it is not supplied by the user, it functions as bias.

\section{Formal Argument}

Let \(x\) be an underdetermined prompt with two possible answers:
\[
S_x=\{a_1,a_2\}.
\]
Assume the premises do not justify preferring either answer:
\[
F_x \not\models a_1 \succ a_2
\quad\text{and}\quad
F_x \not\models a_2 \succ a_1.
\]
A decisive answer requires:
\[
U_{\mathrm{decisive}}(y,x)=1 \Rightarrow y \in \{a_1,a_2\}.
\]
Strict non-bias requires that the model should not select \(a_1\) over \(a_2\), or \(a_2\) over \(a_1\), without a premise-supported ordering:
\[
NB_{\mathrm{strict}}(y,x)=1 \Rightarrow y \notin \{a_1,a_2\}
\]
when no justified preference relation exists. These two requirements conflict:
\[
y \in \{a_1,a_2\} \land y \notin \{a_1,a_2\}.
\]
Furthermore, strong correctness requires entailment:
\[
F_x \models y.
\]
But by assumption:
\[
F_x \not\models a_1
\quad\text{and}\quad
F_x \not\models a_2.
\]
Therefore, neither decisive answer is strongly correct under the original premises. The model can restore correctness only by adding an explicit criterion \(\theta\), producing:
\[
F_x \cup \{\theta\} \models y.
\]
If \(\theta\) is not given by the user or otherwise justified, it becomes an unsupported selection prior. Thus:
\[
D(x) \Rightarrow \neg(C_{\mathrm{strong}} \land NB_{\mathrm{strict}} \land U_{\mathrm{decisive}}).
\]

\section{Bias as a Selection Operator}

For an underdetermined prompt, the model's answer can be represented as an optimization:
\[
y=\arg\max_{a\in S_x} V_{\theta}(a,x),
\]
where \(V_{\theta}\) is a value function and \(\theta\) encodes the model's selection criterion. If \(\theta\) is explicit and user-provided, then the model's answer can be conditionally correct. If \(\theta\) is hidden, then the model's answer may be useful but biased. This is why bias is not merely an accidental defect in decision-oriented generation; in underdetermined settings, some ordering principle is necessary for selecting one answer from many.

\section{Example: Scholarship Selection}

Consider the prompt:
$x = \text{``Student A has GPA 9.5 and moderate financial need. Student B has GPA 8.7 and  }$  
$\text{ severe financial need. Who should receive the scholarship?''}$\\

The premise set is:
\[
F_x=\{GPA(A)=9.5,\;GPA(B)=8.7,\;Need(A)=moderate,\;Need(B)=severe\}.
\]
The possible answers are:
\[
S_x=\{A,B\}.
\]
A merit-based rule gives:
\[
V_{\mathrm{merit}}(i)=GPA(i),
\]
so:
\[
V_{\mathrm{merit}}(A)>V_{\mathrm{merit}}(B)
\Rightarrow y=A.
\]
A need-based rule gives:
\[
V_{\mathrm{need}}(i)=Need(i),
\]
so:
\[
V_{\mathrm{need}}(B)>V_{\mathrm{need}}(A)
\Rightarrow y=B.
\]
A combined rule can be written as:
\[
V_{\alpha}(i)=\alpha \cdot \widehat{GPA}(i)+(1-\alpha)\cdot \widehat{Need}(i),
\quad \alpha \in [0,1],
\]
where \(\widehat{GPA}\) and \(\widehat{Need}\) are normalized scores. If \(\alpha\) is high, the model selects A; if \(\alpha\) is low, it selects B. However, the prompt does not specify \(\alpha\):
\[
F_x \not\models \alpha.
\]
Therefore, a decisive answer requires the model to introduce a criterion. A response such as ``A should be selected because A has the higher GPA'' is correct under a merit-first criterion but biased if merit-first was not specified. A response such as ``B should be selected because B has greater financial need'' is correct under a need-first criterion but biased if need-first was not specified. A response such as ``There is insufficient information; specify whether the scholarship prioritizes merit, need, or a weighted combination'' preserves correctness and non-bias but reduces decisive utility.

\section{Example: Subjective Recommendation}

Consider the prompt:
\[
x=\text{``Which city is better to live in: Bengaluru, Mumbai, or Delhi?''}
\]
The answer space contains at least:
\[
S_x=\{\text{Bengaluru},\text{Mumbai},\text{Delhi},\text{depends on criteria}\}.
\]
A plausible city-ranking function is:
\[
V_{\theta}(c)=w_1Job(c)+w_2Cost(c)+w_3Safety(c)+w_4Climate(c)+w_5Lifestyle(c),
\]
where:
\[
\theta=\{w_1,w_2,w_3,w_4,w_5\}.
\]
A job seeker may place higher weight on \(Job(c)\), a family may place higher weight on \(Safety(c)\), and a student may place higher weight on \(Cost(c)\). If the model says ``Bengaluru is best'', it may be optimizing for technology employment. If it says ``Mumbai is best'', it may be optimizing for finance, entertainment, or metropolitan exposure. If it says ``Delhi is best'', it may be optimizing for policy, administration, or public-sector access. None of these answers is strongly correct unless the relevant weights are supplied. The more faithful answer is conditional: Bengaluru may be preferable under a technology-career criterion, Mumbai under a finance or media criterion, and Delhi under a policy or government-access criterion. Such an answer is less decisive but more transparent.

\section{Example: Normative Decision}

Consider:
\[
x=\text{``Should a company prioritize profit or employee well-being?''}
\]
The possible answers are:
\[
S_x=\{\text{profit},\text{employee well-being},\text{balanced approach}\}.
\]
A decision rule may be:
\[
V_{\theta}(a)=\beta_1FinancialReturn(a)+\beta_2EmployeeWelfare(a)+\beta_3LongTermSustainability(a).
\]
If \(\beta_1\) dominates, the answer favors profit. If \(\beta_2\) dominates, the answer favors employee well-being. If \(\beta_3\) dominates, the answer may favor a long-term balanced strategy. The prompt does not determine \(\beta_1,\beta_2,\beta_3\). Therefore, a singleton answer imports a normative framework. This explains why LLMs often hedge on ethical or social questions: the refusal or conditional answer is not merely evasive; it may be a correctness-preserving strategy.

\section{Refusal as Consistency Preservation}

Let \(r\) denote a refusal or clarification request:
\[
y=r.
\]
For example:
\[
r=\text{``I cannot determine the best candidate without knowing the selection criterion.''}
\]
This response does not assert \(A\) or \(B\), and it does not introduce a hidden preference. Therefore:
\[
C_{\mathrm{strong}}(r,x)=1,\quad NB_{\mathrm{strict}}(r,x)=1,\quad U_{\mathrm{decisive}}(r,x)=0.
\]
This is analogous to a consistency-preserving distributed system under CAP. During partition, the system may refuse a request rather than return an inconsistent answer. During underdetermination, an LLM may refuse or ask for clarification rather than return an unjustified decision.

\section{Managing, Rather Than Eliminating, the Trilemma}

The trilemma is strongest under strict definitions. Practical LLM systems can weaken the conflict by replacing strong correctness with conditional correctness, strict non-bias with transparent assumption disclosure, and decisive utility with assistive utility. Conditional correctness is:
\[
C_{\mathrm{cond}}(y,x,\theta)=1 \iff F_x \cup \{\theta\}\models y.
\]
Transparent non-bias can be approximated by requiring:
\[
Hidden(\theta)=0,
\]
meaning that the model states the criterion used. Assistive utility can be high even when decisive utility is low:
\[
U_{\mathrm{assistive}}(y,x)>0
\quad\text{even if}\quad
U_{\mathrm{decisive}}(y,x)=0.
\]
A response such as ``If the scholarship prioritizes academic merit, select A; if it prioritizes financial need, select B; if both matter, specify a weight \(\alpha\)'' is not maximally decisive, but it is technically stronger than a hidden single-choice recommendation. It exposes the underdetermination rather than concealing it.

\section{Implications for Alignment}

This conjecture has three implications for LLM alignment. First, not every refusal is a failure. Some refusals arise because the prompt asks for a decision that is not entailed by the premises. Second, not every useful answer is unbiased. Usefulness often requires selection, and selection often requires a preference. Third, not every balanced answer is correct. A model that says ``both options are equally good'' may be introducing an unsupported equality claim. These issues matter because foundation models are increasingly used across many downstream tasks, and their failures can propagate across applications \cite{bommasani2021}. They also matter because bias in language systems can reflect both learned social associations and hidden decision criteria \cite{caliskan2017,bender2021}.

A practical design policy is:
\[
\pi(x)=
\begin{cases}
\text{answer directly}, & |S_x|=1,\\
\text{ask for clarification}, & |S_x|>1 \text{ and no criterion is supplied},\\
\text{give conditional answers}, & |S_x|>1 \text{ and several criteria are plausible},\\
\text{recommend with stated assumptions}, & |S_x|>1 \text{ and utility requires a decision}.
\end{cases}
\]
This policy does not eliminate the trilemma. It makes the trade-off explicit.

\section{Conclusion}

This paper proposed a CAP-like trilemma for Large Language Models. The CAP theorem shows that consistency, availability, and partition tolerance cannot all be guaranteed under network partition. Analogously, the proposed correctness--non-bias--utility conjecture states that strong correctness, strict non-bias, and decisive utility cannot always be simultaneously guaranteed under semantic underdetermination:
\[
\boxed{
D(x) \Rightarrow \neg(C_{\mathrm{strong}} \land NB_{\mathrm{strict}} \land U_{\mathrm{decisive}})
}
\]
The central insight is that many LLM decision failures are structurally induced by prompts that request singleton answers without providing singleton-determining premises. In such cases, usefulness requires selection, selection requires a criterion, and an unsupported criterion introduces bias. A model can avoid this bias by refusing, hedging, or asking for clarification, but doing so reduces decisive utility. Therefore, alignment should not only aim to increase correctness, reduce bias, and improve utility independently; it should also study how these desiderata trade off under underdetermination and how models can make those trade-offs explicit.


\vspace{-0.3em}
\begin{thebibliography}{9}
\setlength{\itemsep}{0pt}

\bibitem{brewer2000}
E. A. Brewer, ``Towards robust distributed systems,'' keynote at the ACM Symposium on Principles of Distributed Computing, Portland, Oregon, USA, 2000.

\bibitem{gilbert2002}
S. Gilbert and N. Lynch, ``Brewer's conjecture and the feasibility of consistent, available, partition-tolerant web services,'' \emph{ACM SIGACT News}, vol. 33, no. 2, pp. 51--59, 2002. doi: \href{https://doi.org/10.1145/564585.564601}{10.1145/564585.564601}.

\bibitem{caliskan2017}
A. Caliskan, J. J. Bryson, and A. Narayanan, ``Semantics derived automatically from language corpora contain human-like biases,'' \emph{Science}, vol. 356, no. 6334, pp. 183--186, 2017. doi: \href{https://doi.org/10.1126/science.aal4230}{10.1126/science.aal4230}.

\bibitem{bender2021}
E. M. Bender, T. Gebru, A. McMillan-Major, and S. Shmitchell, ``On the dangers of stochastic parrots: Can language models be too big?,'' in \emph{Proceedings of the 2021 ACM Conference on Fairness, Accountability, and Transparency}, pp. 610--623, 2021. doi: \href{https://doi.org/10.1145/3442188.3445922}{10.1145/3442188.3445922}.

\bibitem{ouyang2022}
L. Ouyang, J. Wu, X. Jiang, D. Almeida, C. L. Wainwright, P. Mishkin, C. Zhang, S. Agarwal, K. Slama, A. Ray, J. Schulman, J. Hilton, F. Kelton, L. Miller, M. Simens, A. Askell, P. Welinder, P. Christiano, J. Leike, and R. Lowe, ``Training language models to follow instructions with human feedback,'' \emph{Advances in Neural Information Processing Systems}, vol. 35, pp. 27730--27744, 2022.

\bibitem{lin2022}
S. Lin, J. Hilton, and O. Evans, ``TruthfulQA: Measuring how models mimic human falsehoods,'' in \emph{Proceedings of the 60th Annual Meeting of the Association for Computational Linguistics}, pp. 3214--3252, Dublin, Ireland, 2022. doi: \href{https://doi.org/10.18653/v1/2022.acl-long.229}{10.18653/v1/2022.acl-long.229}.

\bibitem{bommasani2021}
R. Bommasani et al., ``On the opportunities and risks of foundation models,'' arXiv preprint arXiv:2108.07258, 2021. doi: \href{https://doi.org/10.48550/arXiv.2108.07258}{10.48550/arXiv.2108.07258}.

\end{thebibliography}
\end{document}